\documentclass{llncs}
\usepackage{llncsdoc}

\usepackage{amsmath}
\usepackage{latexsym}
\usepackage{amssymb}

\usepackage{epsf}
\usepackage{lscape}
\usepackage{verbatim}
\usepackage{times}
\usepackage{color}
\usepackage{helvet}
\usepackage{courier}
\usepackage{amsfonts}

\usepackage{multirow}

\usepackage{graphicx}

 \newcommand{\semantics}[1]{[\![ #1 ]\!]} 
\newcommand{\ov}{\overrightarrow} 

\reversemarginpar

\begin{document}
%
\title{Sentence Entailment in\\Compositional Distributional Semantics}
\author{Mehrnoosh Sadrzadeh, Dimitri Kartsaklis, Esma Balk{\i}r
$\{${\it m.sadrzadeh,d.kartsaklis$\}$@qmul.ac.uk}, {\it esmabalkir@gmail.com}}
\institute{School of Electronic Engineering and Computer Science\\
Queen Mary University of London\\
Mile End Road, London E1 4NS, United Kingdom}

\maketitle

\begin{abstract}
Distributional semantic models provide vector representations for words by gathering co-occurrence frequencies from corpora of text. Compositional distributional models extend these  from words to phrases and sentences. In categorical compositional distributional semantics, phrase and sentence representations  are functions of their grammatical structure  and representations of the words therein. In this setting, grammatical structures are formalised by morphisms of a compact closed category and meanings of words are formalised by objects of the same category. These can be instantiated in the form of vectors or density matrices. This paper concerns the applications of this model to  phrase and sentence level entailment. We argue that  entropy-based distances of vectors and density matrices provide a good candidate to measure word-level entailment, show the advantage of density matrices over vectors for word level entailments, and  prove that these distances  extend compositionally from words to phrases and sentences. We exemplify our theoretical constructions on real data and  a toy entailment dataset and provide preliminary experimental evidence.  
\end{abstract}

\noindent
Distributional models of meaning, otherwise known as  distributional semantics,  are based on  the philosophy of  Firth and Harris who argued that meanings of words can be derived from  their patterns of  use   and that words that have similar meanings often occur in the same contexts  \cite{firth1957,Harris}.  For example, words like ``butterfly'' and ``bee'' have similar meanings, since they often occur in the context of ``flower'', whereas ``butterfly'' and ``door'' do not have similar meanings, since one often occurs  close to ``flower'' and one does not.   This hypothesis has been employed to develop semantic vector models  where meanings of words are represented by vectors, built from the frequency of  co-occurrences of words with each other \cite{Rubenstein,Salton}.  Compositional distributional models extend these vector representations from words to phrases and sentences. They  work alongside a principle of compositionality, which states that the meaning of a phrase or sentence is a function of the meanings of the words therein.  Thus, the  vector meaning of  ``yellow butterfly was chased by our cat'', is obtained by acting via a  function, whose form is yet to be decided, on the vector meanings of ``yellow'', ``butterfly'', ``chase'' and ``cat''. Based on how this  function is implemented, these models come in different forms. There are the  ones that use simple point wise vector operations \cite{lapata2010}; these  just add or multiply  vectors of the words. We have  the ones that are based on tensors of grammatical types with vectors of words \cite{ClarkPulman}; these  take the tensor product of the vectors of the words with a vectorial representation of their grammatical types. There are ones where tensors are used to represent meanings of functional words, such as adjectives adverbs, and verbs. Here,  the  functional word gets a tensor meaning and composition becomes tensor contraction \cite{BaroniZam,CoeckeSadrClark2010}. Finally, we have  the ones that use neural word embeddings, where the  function is learnt from data \cite{socher2012,nal2014}.  

The work of this paper is based on the {\em categorical compositional distributional semantics} framework \cite{CoeckeSadrClark2010}, from now on CCDS, where  vectorial meanings of phrases and sentences are built from the vectors and tensors  of the words therein and   the grammatical structures of the phrases and sentences. These models are based on a general mathematical setting, where the meaning of any phrase or sentence, no matter how complex and long they are,  can in principle be assigned a  vectorial representation. Fragments of the model have been instantiated on concrete data and  have been applied to word and phrase/sentence disambiguation, classification, similarity,  and paraphrasing tasks. Some of the instantiations of CCDS in these tasks have  outperformed other compositional distributional  models,  where   for instance, simple operations were used and the grammar was not taken into account, see \cite{grefenstette2011,KartSadrPul-COLING-2013,KartSadr-EMNLP-2013,milajevs2014,GrefenSadrCL}.

In distributional semantics, entailment  is modelled  via the  \emph{distributional inclusion hypothesis}. This hypothesis   says that  word $v$ entails word $w$ when the contexts of  $v$ are included in the contexts of  $w$. This means that whenever word $v$ is used, word $w$ can be used retaining a valid meaning.  The hypothesis  makes intuitive sense, it stands a good chance  for entailment, and indeed there has been an extensive amount of work on it, e.g. see \cite{Dagan1999,weeds2004,kotlerman2010}. However, existing  work is mostly done at the word level and not much has been explored  when it comes to  phrases and sentences. The work  on entailment between quantified noun phrases \cite{baroni2012} is an exception, but it does not take into account  composition and thus does not extend to sentences and longer phrases.  Composition is  what is needed for a modular approach to   entailment and the challenges faced based on it, e.g. see the work described in \cite{dagan2006pascal}. In this and other similar challenges, categorised under the general heading of RTE (Recognising Textual Entailment),  one is to decide about the entailment between complex sentences of language, for example ``yellow butterfly was chased by our cat'' and ``someone's cat chased a butterfly''. In a compositional model of meaning, which is the one we work with, the goal is to try and derive the entailment relation between the sentences from  the entailment relations between the words and the grammatical structures of the sentences.

Two points should be noted here. First  is that entailment is a directional measure, that is if $v$ entails $w$, it is most of the time not the case that $w$ entails $v$. This is in contrast to the notion of similarity, which is computed using symmetric distance measures between vectors, e.g. cosine of the angle, and  is the most common operation in distributional semantics and its applications,  for example see the tasks described in \cite{Schutze,turney2006}. The second point is that, although the distributional inclusion hypothesis can be read in a binary fashion and indeed the notion of entailment in  classical logics has a binary truth value  semantics (i.e. either it holds or not), in a distributional setting it would make more sense to work with degrees of entailment.  Conceptually, this is because we are in a quantitative setting that represents meanings of words by vectors of numbers rather than in the qualitative setting of classical logic,  designed to reason about truth valued predicates.  Concretely and when it comes to working with data, it is rarely the case that one gets 0's in the coordinates of vectors. Some coordinates might have low numbers; these should  be used in a lesser extent in the entailment decision. Some coordinates have large numbers; these should affect the entailment decision to a larger extent. In summary, in order to model entailment in a distributional semantics one is after an operation between the vectors  that is asymmetric (similar to the logical entailment) and has degrees (contrary to the logical entailment).   This is exactly what previous work on word-level entailment \cite{Dagan1999,weeds2004,kotlerman2010} has done and what we are going to do in this paper for phrase/sentence-level entailment. 

In this paper we show how CCDS can be used to reason about entailment in a compositional fashion. In particular, we prove how the general compositional procedures of this model give rise to  an entailment relation at the word level  which is extendible to the phrase and sentence level. At the word level, we work with the distributional inclusion hypothesis.   Previous work on word level entailment in these models shows that entropy-based notions such as KL-divergence provide a good notion of degrees of entailment based on the distributional inclusion hypothesis \cite{Dagan1999,herbelot2013,Rimell14}. In this paper, we prove that in CCDS this notion  extends  from word vectors to phrase and sentence vectors and thus also provides a good notion of phrase/sentence entailment: one that is  similar to that of Natural Logic \cite{MacCartney2007}.  We also show that in the presence of correlations between contexts, the notion of   KL-divergence  naturally lifts from vectors to density matrices via von Neumann's entropy, and that this notion of entropy also lifts compositionally from words to phrases and sentences,  in the same way as KL-divergence did for vectors.  

The density matrix results of this paper  build on the  developments of \cite{balkir2014,Balkir15} and are related to \cite{piedeleu2015,Piedeleu-Msc,kartsaklisphd}, where the use of density matrices in CCDS were initiated. More recently, the work of \cite{Martha16} focuses on the density matrices of CCDS  to develop a theoretical notion for a  \emph{graded entailment} operator. Prior to that, density matrices were  used in \cite{blacoe2013} to assist in parsing.   
In contrast to these works,  here (and in the conference version of this paper \cite{ISAIM16}), we do not start right away with density matrices, neither do we treat density matrices  as our only or  first-class citizens. The main contribution of our work is that we develop a more general notion of entailment that is applicable to both vectors and density matrices.  This notion is compositional and extends modularly from words to phrases and sentences. The reason for the fact that our results hold  for both vectors and density matrices is that they  are both instances of the same higher order categorical structure: the category of vector spaces and linear maps and the category of density matrices and completely positive maps are both compact closed.

The outline of our contribution is as follows. We start  with vectors and vector-based notions of entropy,  pointing out a shortcoming of vector-level entropy when  it comes to measuring a certain form of entailment,   motivate  how this problem can be solved using  density matrices, and then move on to show how one can incorporate in  the CCDS setting an entailment based on density matrices. In short, we develop a  distributional notion of entailment that  extends compositionally from words to phrase and sentences and which works for both vectors and density matrices. We     argue, in theoretical,  in concrete, and in experimental  terms, that the notion of relative entropy on density matrices gives rise to a richer notion of word and sentence level entailment than the notion of KL-divergence on vectors.

On the concrete side, we provide two small scale experiments on data collected from a text corpus, build vectors and density matrices, and  apply the results  to a toy  word level entailment task and a short phrase and sentence  entailment task. This involves  implementing a concrete way of  building  vectors and density matrices for words and composing them to obtain vectors and density matrices for our short sentences. We elaborate on all of these in the corresponding  sections of the paper.  As will be pointed out below, some of the concrete constructions we present are novel.  

This paper  is the journal version of the work  presented in the 14th International Symposium in Artificial Intelligence and Mathematics \cite{ISAIM16}. The novel contributions of the current paper, in relation to its conference version,  are as follows:

\begin{enumerate}
\item We prove a more general version of the main result of the previous paper, i.e. Theorem \ref{thm:main}. In the new version,  this theorem is not restricted to sentences that satisfy Lambek's switching lemma, which says that the grammatical structures of sentences are only  epsilon maps (i.e. applications of functions) and identities. Here, we show that the grammatical structures of phrases/sentences can be any morphism of a base  category of finite dimensional vector spaces (for the vectorial entailments) and a base category of density matrices and completely positive maps (for the density matrix entailments).


\item We develop and implement a new way of building concrete density matrices for words, thus work with two different concrete implementations, as opposed to the only one presented in the conference version. In the previous method, a density matrix was created as a convex combination of vectors representing contexts, following the quantum-mechanical intuition. The new method, on the other hand,  is  based on the philosophy that  there might exist  correlations between the contexts, and it directly implements the reasoning presented in Section \ref{sec:dih}.  The examples of that section  argue in favour of density matrices over vectors for basis correlation cases, and our new density  matrices are built in  the same way as prescribed by the general pattern present in  such  cases.  

\item We present additional analysis based on a new toy example for cases where there is a  correlation between the  contexts (in other words basis vectors/words), and show that density matrices built using the method described above do respect  the entailment relations in these cases and do so better than vectors. 

\item  Finally, we take advantage of the space provided in the journal version and provide more background on the categorical constructions used in CCDS. 

\end{enumerate} 

\section{Categorical Preliminaries and Examples}
\label{sec:categoryprelim}

Categorical Compositional Distributional Semantic (CCDS)  relies on the theory of categories, originated in the work of MacLane \cite{CWM}. It is based on a special type of categories, known as compact closed categories,  developed in \cite{KellyLaplaza80}. We   will briefly  recall a few of the  major notions that are important to our work  from these theories and refer the reader for the complete list of definitions and properties to  \cite{CWM,KellyLaplaza80}. An introduction to the subject with a focus on compact closed categories  is presented in \cite{coeckepaquette}.

The main inhabitants of a  category ${\cal C}$ are its \emph{objects} and \emph{morphisms}. The objects are denoted by  $A, B, C$  and the morphisms by  $f,g$. If $f$ is a morphism from $A$ to $B$, we denote it by   $f \colon A \to B$,  similarly $g \colon B \to C$ denotes  a morphism from $B$ to $C$.  Each object $A$ has an identity morphism, denoted by $1_A \colon A \to A$.  The morphisms are closed under composition, that is,  given $f \colon A \to B$ and $g \colon B \to C$, there is a morphism $g \circ f \colon A \to C$ from $A$ to $C$. Composition is associative, that is:
 \[
 f\circ (g \circ h) = (f \circ g) \circ h
 \]
 with identity morphisms  its  units, that is:
   \[
   f  \circ 1_A = f \qquad \text{and}  \qquad 1_B \circ f = f
   \]
A monoidal  category has  a binary operation  defined on its objects and morphisms, referred to as \emph{tensor} and denoted by $A \otimes B$ on objects and similarly by $f \otimes g$ on morphisms. This operation  is associative and has a unit $I$, which is an object of the category. Associativity of  tensor and it having a unit means that we have:
\[
A \otimes (B \otimes C) = (A \otimes B) \otimes C \qquad 
A \otimes I = I \otimes A = A
\]
On a  pair of morphisms $(f:A\to C, g: B\to D)$, the tensor operation is defined as  follows:
\[
f\otimes g:A\otimes B\to C\otimes D
\] 
 It satisfies  a \em bifunctoriality \em  property, that is, the following equation holds:
\[
(g_1\otimes g_2)\circ(f_1\otimes f_2)=(g_1\circ f_1)\otimes (g_2\circ f_2)\,.
\]
for $f_1, f_2 \colon A \to C$ and $g_1, g_2\colon B \to D$. 

A compact closed category  is a monoidal category,  where  
each of its  objects has  two  contravariant functors defined on them; these are  referred to as  \emph{left and right adjoints} and they are to satisfy an \emph{adjunction} property.  Given an object $A$, its adjoints are denoted by $A^r$ and $A^l$ and are referred to as \emph{right} and \emph{left} adjoints. Part of the  property they satisfy says that they are equipped  with the following  morphisms:
\begin{align*}
A \otimes A^r \stackrel{\epsilon_A^r} {\longrightarrow} \; &I
\stackrel{\eta_A^r}{\longrightarrow} A^r \otimes A \hspace{1cm}
A^l \otimes A \stackrel{\epsilon_A^l}{\longrightarrow} \; I
\stackrel{\eta_A^l}{\longrightarrow} A \otimes A^l\
\end{align*}
In other words, for each object $A$, there exists in a compact closed category  an object $A^r$, an object $A^l$ and the above four morphisms. 
These morphisms satisfy the following equalities, sometimes referred to by the term \emph{yanking}:
\begin{align*}
& (1_A \otimes \epsilon_A^l) \circ (\eta_A^l \otimes 1_A)  = 1_A 
\hspace{1cm}
&&(\epsilon_A^r \otimes 1_A) \circ (1_A \otimes
  \eta_A^r)   = 1_A\\
& (\epsilon_A^l \otimes 1_{A^l}) \circ (1_{A^l} \otimes
  \eta_A^l) = 1_{A^l}  
    \hspace{1cm}
&&    (1_{A^r} \otimes \epsilon_A^r) \circ (\eta_A^r \otimes 1_{A^r}) = 1_{A^r}
\end{align*}

A self adjoint compact closed category is one in which the objects and their adjoints are the same, that is for every object $A$ we have 
\[
A^l = A^r = A
\]

A  \emph{strongly monoidal functor} $F$ between a  monoidal  category ${\cal C}$ and another monoidal category ${\cal D}$ is a map $F \colon {\cal C} \to {\cal D}$, which  assigns to each object $A$ of ${\cal C}$ an object $F(A)$ of ${\cal D}$ and to each morphism  $f \colon A \to B$ of ${\cal C}$, a morphism $F(f) \colon F(A) \to F(B)$ of ${\cal D}$.  It  preserves the identities and the compositions of ${\cal C}$. That is, we have 
\[
F(1_A) = 1_{F(A)}
\qquad 
F(g \circ f) = F(g) \circ F(f)
\]
Moreover, we have the following equations:
\[
F(A \otimes B) = F(A) \otimes F(B) \qquad
F(I) = I
\]
These mean that  $F$ preserves the tensor and its unit in both directions. A strongly monoidal functor on  two  compact closed categories ${\cal C}$ and ${\cal D}$ preserves the adjoints, that is we have:
\[
F(A^l) = F(A)^l \qquad
F(A^r) = F(A)^r
\]

The above definitions are given in a  \emph{strict} monoidal sense. In a non-strict setting, the equalities of the monoidal properties are replaced with isomorphisms. We work with three examples of compact closed categories: pregroup algebras,  the category of finite dimensional vector spaces and linear maps, and the completely positive maps over them. Below we show how each of these is a compact closed category. 

\paragraph{\bf Pregroup algebras \textmd{PRG}.}

A pregroup algebra is   a partially ordered monoid where each element has a left and a right adjoint; it is denoted by  $\textmd{PRG} = (P, \leq, \cdot, 1, (-)^r, (-)^l)$. The notion of adjunction here means that for each $p \in P$, we have a $p^r$ and a $p^l$ in $P$ such that:
 \[
 p \cdot p^r \leq 1 \leq p^r \cdot p \qquad
 p^l \cdot p \leq 1 \leq p \cdot p^l
 \]
 A pregroup algebra is a compact closed category in the following way: the elements of the partial order $p \in P$ are the objects of the category. The partial orderings between the elements are the morphisms of the category, that is for $p,q \in P$ we have:
 \[
 p \to q \qquad \text{iff} \qquad p \leq q
 \]
The monoid multiplication of the pregroup algebra is a monoidal tensor; this is because we can form the monoid multiplication of elements of the partial ordering $p \otimes q$ and that this multiplication preserves the ordering, that is we have:
\[
p \leq  q \quad \text{and} \quad p' \leq q' \ \implies \ p \otimes p' \leq q \otimes q'
\]
The unit of this multiplication is 1, since we have:
\[
p \cdot 1 = 1 \cdot p = p
\]
The multiplication is associative as well, as denoted via the following inequality which holds in $\textmd{PRG}$:
\[
p \cdot (q \cdot r) = (p \cdot q) \cdot r
\]
Each element of the pregroup algebra has a left and a right adjoint and the adjunction inequalities expressed above mean that the adjunction morphisms exist, that is we have the following:
\[
p \otimes  p^r  \stackrel{\epsilon_p^r}{\to} 1 \stackrel{\eta_p^r}{\to} p^r \otimes p \qquad
 p^l \otimes p \stackrel{\epsilon_p^l}{\to} 1 \stackrel{\eta_p^l}{\to}  p \otimes p^l
\]
In order to see that the above satisfy the  yanking equalities, consider the first  yanking case, which is as follows:
\[
(1_A \otimes \epsilon_A^l) \circ (\eta_A^l \otimes 1_A)  = 1_A 
\]
In a pregroup algebra setting, this will look like as follows:
\[
(1_p \otimes \epsilon_p^l) \circ (\eta_p^l \otimes 1_p)  = 1_p
\] 
We form $(\eta_p^l \otimes 1_p)$ by multiplying both sides of the $\eta_p^l$ inequality by $p$: 
\[
1 \leq p \cdot p^l  \ \implies \ 1 \cdot p \leq p \cdot p^l \cdot p
\]
Similarly, we form $(1_p \otimes \epsilon_p^l)$ by multiplying both sides of the $\epsilon_p^l$ inequality by $p$: 
\[
p^l \cdot p \leq 1  \ \implies \ p \cdot p^l \cdot p \leq p \cdot 1 
\]
Then we compose these two morphisms, which in partial order terms amounts to applying the transitivity of the partial order to them, as follows
\[
1 \cdot p \leq p \cdot p^l \cdot p \leq p \cdot 1 
\] 
Thus we  obtain the following inequality:
\[
1 \cdot p \leq 1 \cdot p
\] 
which is true since  the partial order is reflexive. The other three yanking equalities are proven  in the same  way. 

\paragraph{\bf Finite-dimensional vector spaces with fixed orthonormal basis and linear maps.}

Finite dimensional vector spaces  over reals $\mathbb{R}$ and the linear maps between the spaces form a compact closed category, denoted by $\textmd{FVect}_{\mathbb{R}}$. The objects of this category are the vector spaces, while its morphisms are the linear maps between them.   The monoidal tensor of the category is the tensor product of  vector spaces which can be extended to linear maps as follows: For two linear maps $V \stackrel{f}{\to} W$ and $V' \stackrel{g}{\to} W'$, their tensor is denoted by $f \otimes g$ and  is defined to be the  following map:
\[
V \otimes V' \stackrel{f \otimes g} {\longrightarrow} W \otimes W'
\] 
The unit of the monoidal tensor is the unit of the tensor product of the vector spaces, which is the scalar field, since we have the following for every vector space $V$:
\[
V \otimes \mathbb{R} \cong \mathbb{R} \otimes V \cong V
\]
For each vector space $V$, its dual space $V^*$ is  its left and right adjoint, that is:
\[
V^l = V^r := V^*
\]
In the presence of a fixed orthonormal basis, which is the case here and for vector spaces of a distributional semantics, we have a way of transforming $V^*$ to $V$ and $V$ to $V^*$. Such categories, denoted by  $\textmd{FdVect}_{\mathbb{R}}$, are thus self adjoint compact closed categories.   Moreover, their tensor (and the tensor of of $\textmd{FVect}_{\mathbb{R}}$ more generally)  is symmetric, that is we have:
\[
V \otimes W \cong W \otimes V
\]
As a result, the two $\epsilon^r$ and $\epsilon^l$ maps become the same map and similarly so for the $\eta$ maps. That is we have:
\[
 \epsilon := \epsilon^r  \cong \epsilon^l  \qquad
 \eta := \eta^r \cong \eta^l 
\]
Thus the $\epsilon$ and $\eta$ maps of this category will acquire the following forms:
 \[
 \epsilon_V \colon V \otimes V \to \mathbb{R}\qquad \qquad
 \eta_V \colon \mathbb{R} \to V \otimes V
 \]
Given  $\sum_{ij} C_{ij} \overrightarrow{v_i} \otimes \overrightarrow{v_j} \in V\otimes V$ and a basis   $\{\overrightarrow{v}_i\}_i$ for $V$, the above  are concretely defined as follows:
\[
\epsilon_V \left(\sum_{ij} C_{ij} \overrightarrow{v_i} \otimes \overrightarrow{v_j}\right) := \sum_{ij} C_{ij}  \langle \overrightarrow{v_i} | \overrightarrow{v_j} \rangle
\]
for the $\epsilon$ map and as follows:
\[
\eta(1) := \sum_i \overrightarrow{v_i} \otimes \overrightarrow{v_i}
\]
for the $\eta$ map.  In order to see that the above satisfy the yanking equalities, again consider the first yanking equality; in its vectorial form, for one side of the equality we have to build the following morphism:
\[
(1_V \otimes \epsilon_V) \circ (\eta_V \otimes 1_V) 
\]
which is  obtained by the following composition of morphisms:
\[
\mathbb{R} \otimes V \stackrel{\eta_V \otimes 1_V}{\longrightarrow} V \otimes V \otimes V 
\stackrel{1_V \otimes \epsilon_V}{\longrightarrow} V \otimes \mathbb{R}
\]
This is equal to the identity morphism on $V$,  since we have:
\[
\mathbb{R} \otimes V \cong V \otimes \mathbb{R}  \cong V
\]
 due to the fact that $\mathbb{R}$ is the unit of tensor in $\textmd{FVect}$.

\paragraph{\bf Finite-dimensional vector spaces and completely positive maps ${\cal CPM}(\textmd{FVect}_{\mathbb{R}})$.}
 
The category ${\cal CPM}(\textmd{FVect}_{\mathbb{R}})$ over finite dimensional vector spaces and linear maps  is also compact closed. The ${\cal CPM}$ construction was originally defined over Hilbert spaces \cite{selinger2007}. In previous work, we show how it also applies to the simpler case of vector spaces over reals \cite{Balkir15}. The corresponding construction yields a category whose objects are of the form  $V \otimes V^*$, elements of which represent density operators. This property is referred to by the Choi-Jamiolkowski correspondence,  for more on this see \cite{coeckepaquette}. Recall that these are self-adjoint,  semi-definite positive, and have trace 1.  The general form of a density matrice $\hat{v} \in V \otimes V^*$ is as follows:
\begin{equation}
\label{eq:dm-qm}
\hat{v} := \sum_{i} p_i \ov{c}_i \otimes \ov{c}_i
\end{equation}
where $p_i$'s  define a probability distribution over the set of  $\ov{c}_i$ vectors, thus we have:
\[
0\leq p_i \leq 1 \qquad\qquad \sum_i p_i = 1
\]
The $\ov{c}_i$ vectors are referred to by \emph{pure} states and the $\hat{v}$ is referred to by a \emph{mixed} state, in quantum mechanic terminology. 

Morphisms of ${\cal CPM}(\textmd{FVect}_{\mathbb{R}})$ are linear maps which are moreover completely positive. Again, recall that a completely positive map between two density matrices preserves the structure of a density matrix.  In category theoretic terms, these maps are  morphisms of the following form:
\[
f \colon V \otimes V^* \to W \otimes W^*
\]
for which there exist a vector space $X$ and a linear map $g \colon V \to X \otimes W$ such that the following map exists in $\textmd{FVect}_{\mathbb{R}}$:
\[
f = (g \otimes g) \circ (1_{W \otimes W} \otimes \eta_{X})
\]
The category ${\cal CPM}(\textmd{FVect}_{\mathbb{R}})$ inherits the symmetry property of $\textmd{FVect}_{\mathbb{R}}$, that is we have:
\[
(V \otimes V^*) \otimes (W \otimes W^*)  \cong (W \otimes W^*) \otimes (V \otimes V^*)
\]
Also, similar to $\textmd{FVect}_{\mathbb{R}}$, its  left and right adjoints become equal and reduce to the tensor product of dual spaces. This   is easily shown as follows for the left adjoint and by recalling that  $(-)^*$ is involutive and that the compact closure  is self adjoint:
\[
(V \otimes V^*)^l = (V^*)^l (\otimes)^* V^* \cong (V^*)^* \otimes V^* \cong  V \otimes V^*
\]
The case of right adjoint is similar.   Also, similar to $\textmd{FVect}_{\mathbb{R}}$, in the presence of a fixed basis, the category becomes self adjoint, that is we  have:
\[
(V \otimes V^*)^* \cong V \otimes V^*
\]
The  $\epsilon$ and $\eta$ maps of ${\cal CPM}(\textmd{FVect}_{\mathbb{R}})$ are  obtained by tensoring the $\epsilon$ and $\eta$  maps in $\textmd{FVect}_{\mathbb{R}}$. In the presence of a fix basis, these will have the following forms:
\[
\epsilon \colon V \otimes V \otimes V \otimes V \to \mathbb{R}
\qquad
\eta \colon \mathbb{R} \to V \otimes V \otimes V \otimes V 
\]
Concretely, these maps are given as follows for the $\epsilon$ case:
\[
\epsilon_V \left(\sum_{ijkl} C_{ijkl} \overrightarrow{v_i} \otimes \overrightarrow{v_j} \otimes  \overrightarrow{v_k} \otimes \overrightarrow{v_l}\right) := \sum_{ijkl} C_{ijkl}  \langle \overrightarrow{v_i} | \overrightarrow{v_j} \rangle \langle \overrightarrow{v_k} | \overrightarrow{v_l} \rangle
\]
and as follows for the $\eta$ case:
\[
\eta(1) := \sum_{i} \overrightarrow{v_i} \otimes \overrightarrow{v_i} \otimes \overrightarrow{v_i} \otimes \overrightarrow{v_i}
\]  
Finally, we leave it to the reader to verify that the yanking equalities are also satisfied in a very similar way  they are satisfied in $\textmd{FVect}_{\mathbb{R}}$. 

\section{Categorical Compositional Distributional Semantics (CCDS)}

 In its most abstract form, a CCDS is denoted as follows:
 \[
 ({\cal C}_{\textmd{Syn}}, {\cal C}_{\textmd{Sem}}, F,  \semantics{\ })\]
It  consists of a compact closed category for syntax ${\cal C}_{\textmd{Syn}}$, a compact closed category for semantics ${\cal C}_{\textmd{Sem}}$,  a strongly monoidal functor $F\colon {\cal C}_{\textmd{Syn}} \to {\cal C}_{\textmd{Sem}}$ between the two,  and a semantic map $\semantics {\ } \colon \Sigma^* \to {\cal C}_{\textmd{Sem}}$ from the set of strings of a language $\Sigma^*$ to the compact closed category of semantics. 

Meanings of phrases and sentences of a language are related to the meanings of words of that language via a principle known to the formal semanticist as the principle of \emph{lexical substitution}.  In a CCDS, this principle  takes the following form:
 \begin{equation}
 \label{eq:lexsub}
 \semantics{w_1w_2 \cdots w_n} := F(\alpha)(\semantics{w_1} \otimes \semantics{w_2} \otimes \cdots \semantics{w_n})
 \end{equation}
 for $w_1 w_2 \cdots w_n \in \Sigma^*$ a string of words, i.e. we have  $w_i \in \Sigma$  for each $w_i$ in the string, and where $\alpha$ denotes the grammatical structure of $w_1 w_2 \cdots w_n$, i.e. a morphism in the compact closed category of syntax ${\cal C}_{\textmd{Syn}}$. On the left-hand side of the above equation,  $\semantics{w_1 w_2 \cdots w_n}$ is the  semantics of a string of words and on the right-hand side, each $\semantics{w_i}$ is the semantics of a word in that string.  
 
In practice, the abstract model is instantiated to concrete settings.  One needs a concrete setting to represent the syntax, a concrete setting to represent the semantics,   a concrete way of relating the words of a language, i.e. elements of $\Sigma$,  to semantic representations in ${\cal C}_{\textmd{Sem}}$, and a concrete way of relating the syntactic elements to their semantic counterparts, that is a concrete way of representing the functor $F$ on atomic elements of syntax and semantics. Below, we show how one can do such a many-fold instantiation  for the  cases of $\textmd{PRG}$ for syntax and  $\textmd{FVect}_{\mathbb{R}}$ for vector semantics, and for the cases of $\textmd{PRG}$ as syntax and  ${\cal CPM}(\textmd{FVect}_{\mathbb{R}})$ for density matrix semantics. 

 
\subsection{Instantiation to $(\textmd{PRG}, \textmd{FVect}_{\mathbb{R}}, F, \semantics{\ })$}

In this instantiation, on the syntactic side, we work with a pregroup grammar; this is a pregroup algebra applied to reasoning about syntax and grammatical structures and has been developed by Lambek \cite{lambek2001}.  We provide an overview below.

A pregroup grammar is a pregroup algebra  denoted by  $T(B)$; this notation is to express the fact that the pregroup algebra is  generated over the set $B$ of basic grammatical types of a language. We assume $B$ to be  the set $\{n, s\}$, where $n$ denotes the type of a noun phrase and $s$ the type of a sentence. The pregroup grammar comes equipped with a relation $R \subseteq T(B) \times \Sigma$ that assigns grammatical types from $T(B)$ to the vocabulary $\Sigma$ of a language. Some examples from the  English language are as follows:
\begin{center}
\begin{tabular}{c|c||c}
Grammatical Relation & Pregroup Type & Examples\\
\hline
\hline
adjectives  & $n \cdot n^l$ & red, big, round\\
intransitive verbs & $n^r \cdot s$ & sleep, sneeze, snooze \\
 transitive verbs & $n^r \cdot s \cdot n^l$ & gave, hold, own\\
 adverbs & $s^r \cdot s$ &  yesterday, quickly, slowly
 \end{tabular}
 \end{center}
 
 In a pregroup grammar,   the grammatical structure of a string of words  $w_1 w_2 \cdots w_n$, for $w_i \in \Sigma$,  is the following morphism of  category  $\textmd{PRG}$:
 \[
 t_1 \cdot t_2 \cdot \cdots \cdot t_n \stackrel{\alpha}{\to} t
 \]
where we are taking  $\textmd{PRG}$ to be the compact closed categorical form of our pregroup algebra $T(B)$.  Each $t_i$ is a grammatical  type assigned to the word $w_i$. Formally, this means that we have $t_i \in R[w_i]$. By means of examples, each $t_i$ lives in the middle column of the exemplary table above. For example, for a word $w_5 = $ `red',  we have that $t_5 = n \cdot n^l$, for $w_{18} = $ `sleep', we have that $t_{18} = n^r \cdot s$, and so on.    

 On the semantic side, we work with  $\textmd{FVect}_{\mathbb{R}}$, as previously introduced, that is the compact closed category of finite dimensional vector spaces and linear maps. Thus, our  syntax-semantics map is  a strongly monoidal functor with the following form:
\[
F \colon  \textmd{PRG} \to \textmd{FVect}_{\mathbb{R}}
\]
The concrete form of the functor we are interested  in  acts as follows on the  basic types  of $\textmd{PRG}$:

\[
F(n) := N \qquad F(s) = S
\]
where  $N$ and $S$ are two vector spaces in $\textmd{FVect}_{\mathbb{R}}$. The  strong monoidality of $F$  results in  certain   equalities on the non-atomic elements of $\textmd{PRG}$, examples of which are as follows:
\[
F(p \cdot q ) = F(p) \otimes F(q)\quad
F(1) = \mathbb{R} \quad
F(p^r) = F(p^l) =F(p)^*
\]
These extend to the morphisms, for example we have the following morphism inequalities:
\[
F(p\leq q) = F(p) \to F(q)
\qquad 
F(p \cdot p^r \leq 1) = \epsilon_{F(p)}\qquad
F(1 \leq p^r \cdot p) = \eta_{F(p)}
\]
as well as the following  similar ones for the left adjoints:
\[
F(p^l \cdot p \leq 1) = \epsilon_{F(p)}\qquad
F(1 \leq p\cdot p^l) = \eta_{F(p)}
\]

In this setting, the  meaning representations  of words  are vectors; that is, $\semantics{v}$, for $v$ a word or a string of words, is  a vector $\ov{v}$, hence the principle of  lexical substitution instantiates as follows:
\begin{equation}
\label{vectlexsub}
\overrightarrow{w_1 w_2 \cdots w_n} := F(\alpha) (\overrightarrow{w}_1 \otimes \overrightarrow{w}_2 \otimes \cdots \otimes \overrightarrow{w}_n)
\end{equation}
for $\overrightarrow{w_1 w_2 \cdots w_n} $ the vector representation of the string $w_1 w_2 \cdots w_n$ and $\ov{w_i}$ the vector representation  of word $w_i$ in the string. 

\subsection{Instantiation to $(\textmd{PRG}, {\cal CPM}(\textmd{FVect}_{\mathbb{R}}) , F, \semantics{\ })$}

The syntactic side is as in the previous case. On the semantic side,  we work in the  compact closed category ${\cal CPM}(\textmd{FVect}_{\mathbb{R}})$. The passage from  $\textmd{FVect}_{\mathbb{R}}$ to ${\cal CPM}(\textmd{FVect}_{\mathbb{R}})$ is functorial. Thus, the categorical compositional distributional semantics works along the following functor:
\[
F \colon \textmd{PRG} \to \textmd{FVect}_{\mathbb{R}} \to {\cal CPM}(\textmd{FVect}_{\mathbb{R}})
\]
Here, the  meaning representations  of words are density matrices, that is $\semantics{v}$ is $\hat{v}$, for $v$ a word or a string of words, hence the  principle of  lexical substitution  instantiates as follows:
\begin{equation}
\label{denstlexsub}
\widehat{w_1 \cdots w_n} := F(\alpha)(\hat{w}_1 \otimes \cdots \otimes \hat{w}_n)
\end{equation}
for  $\widehat{w_1 \cdots w_n}$ the density matrix representation of the string $w_1 w_2 \cdots w_n$ and $\hat{w}_i$ the density matrix representation of word $w_i$, for each word of the string. 

\section{KL-Divergence and Relative Entropy}
For a vector space $V$ with a chosen orthonormal basis $\{\ov{v_i}\}_i$, a normalized vector $\ov{v} = \sum_i p_i \ov{v_i}$ can be seen as a probability distribution over the basis. In this case one can define a notion of entropy for $\ov{v}$ as follows:
\[
S(\ov{v}) = -  \sum_i p_i \ln p_i
\]
which is the same as the entropy of the  probability distribution $P = \sum_i p_i$ over the basis.  

For two vectors $\ov{v}, \ov{w}$ with probability distributions $P$ and $Q$, the distance between their entropies, referred to by  Kullback-Leibler divergence, is  defined as:
\[
KL (\ov{v} \| \ov{w}) = \sum_j p_j (\ln p_j - \ln q_j)
\]
This is a measure of distinguishability. One can define a degree of representativeness based on this measure: 
\[
R_{KL}(\ov{v}, \ov{w}) = \frac{1}{1 + KL(\ov{v} \|\ov{w})}
\] 
This is a real number in the unit interval. When  there are non zero weights on the basis elements of $\ov{v}$ that are zero in $\ov{w}$, then $\ln 0 = \infty$ (by convention $0 \ln 0 = 0$) and so $R_{KL}(\ov{v}, \ov{w}) = 0$.  So when the support of  $P$ is not included in the support of $Q$ then $R_{KL} = 0$, and when $P=Q$ then $R_{KL} = 1$. 

Both KL-divergence and representativeness are asymmetric measures. The following measure, referred to by Jensen-Shannon divergence, provides a symmetric version:
\[
JS (\ov{v}, \ov{w}) = \frac{1}{2} \left[ KL\left(P \| \frac{P + Q}{2} \right)  \ + \  KL\left(Q \| \frac{P + Q}{2}\right) \right]
\]

If there are correlations between  the  basis   of $V$, these can be represented by a positive semi-definite symmetric matrix. Suppose we write this matrix in the chosen orthonormal basis as \hbox{$\hat{v} = \sum_{ij} p_{ij} \ov{v_i} \otimes \ov{v_j}$}. The diagonal entries of $\hat{v}$ are  probabilities over the basis, so we have: 
\[
\sum_{ii} p_{ii} = 1
\]
The non-diagonal entries denote the correlations between the basis.   The correlation between $\ov{v_i}$ and $\ov{v_j}$ is the same as the correlation between $\ov{v_j}$ and $\ov{v_i}$.  The matrix $\hat{v}$ given in the form above is the matrix form of a density operator in the chosen basis $\{\ov{v_i}\}_i$.



Density matrices have a notion of entropy, called von Neumann entropy, defined as follows:
\[
N(\hat{v}) = - \textmd{Tr} (\hat{v} \ln \hat{v})
\]
They also have a notion of  KL-divergence:
\[
N(\hat{v} || \hat{w}) = \textmd{Tr}\  \hat{v} (\ln \hat{v} - \ln \hat{w})
\]
The representativeness between two density matrices is  defined in a similar way as for vectors. 
It is a real number in the unit interval, with 0 and 1 values as described before:
\[
{R}_N(\hat{v}, \hat{w}) = \frac{1}{1 + N(P ||Q)}
\]
The density matrix version of the Jensen-Shannon divergence is obtained by replacing $S$ with $N$. 

 A vector can be represented as a diagonal density matrix on the chosen basis $\{\ov{v_i}\}_i$.  In this case, entropy and von Neumann entropy are the same, since the density matrix has no information on its non-diagonal elements, denoting a zero correlation between the chosen basis.

\section{Distributional Inclusion Hypothesis for Vectors and Density Matrices}
\label{sec:dih}

According to the distributional inclusion hypothesis (DIH)  if word $v$ entails word $w$ then the set of  contexts of $v$ are included in the set of  contexts of $w$. This makes sense since it means that whenever word $v$ is used in a context, it can be replaced with word $w$,  in a way such that the meaning of $w$ subsumes  the meaning of $v$. For example,  `cat' entails `animal', hence in the sentence `A cat is drinking milk', one can replace `cat' with `animal' and the meaning of the resulting sentence subsumes that of the original sentence.  On the other hand, `cat' does not entail `goldfish', evident from the fact that the sentence `A goldfish is drinking milk' is very unlikely to appear in a real corpus. 

Different asymmetric measures on probability distributions have been  used to model   and  empirically evaluate the DIH.  Entropy-based measures such as KL-divergence is among successful such  measures. Take the orthonormal basis of a distributional space to be the context lemmas of a corpus and this measure becomes zero if there are contexts with zero weights  in $\ov{v}$ that do not have zero weights in $\ov{w}$. In other words, $R_{KL}(\overrightarrow{v}, \overrightarrow{w}) = 0$ when $v$ does not entail $w$.  The contrapositive of this provides a degree of entailment:
\begin{equation}\label{vectentail}
\overrightarrow{v} \vdash \overrightarrow{w} 
 \quad \Rightarrow
\quad
R_{KL}(\overrightarrow{v}, \overrightarrow{w}) \neq 0
\end{equation}
The $\alpha$-skew divergence  of Lee \cite{Lee1999} and a symmetric version of it based on $JS$ \cite{Dagan1999} are variations on the above. 

Similarly, for density matrices one can use the degree of representativeness of two density matrices  $R_N$ to check for inclusion of contexts. 
\begin{equation}\label{densityentail}
\hat{v} \vdash \hat{w} 
 \quad \Rightarrow
\quad
{R}_N(\hat{v}, \hat{w}) \neq 0
\end{equation}
Here contexts can be single context lemmas for the diagonal elements where the basis are reflexive pairs $(p_i, p_i)$; contexts can also be pairs of two context lemmas for the non-diagonal elements where the basis are pairs $(p_i, q_j)$ with $p_i \neq q_j$. Hence, not only we are checking inclusion over  single contexts,  but also over correlated contexts. The following example shows why this notion  leads to a richer notion of entailment.

\begin{example}
\label{eg:goldfish}
For the sake of simplicity suppose  we do not care about the frequencies per se, but  whether the bases occurred with the target word at all. So the entries are always either 1 or 0. Consider a distributional space with basis $\{$aquarium, pet, fish$\}$ and two target words: `cat' and `goldfish' therein. Assume that we have seen `cat' in the context of  `fish', and also independently, in the context of `pet'. Assume further that we have seen the word `goldfish' in the context of  `aquarium', and also in the contexts of  `pet' and `fish', but whenever  it was in the context of `pet', `fish' was also around: for example they always occurred in the same sentence. Hence, we have never seen `goldfish' with `pet' or `fish' separately. This signifies a correlation between `pet' and `fish' for the target word `goldfish'. 

This correlation is not representable in the  vector case and as a result,  whereas  `cat' does not  normally entail `goldfish', its vector representation does, as the set of contexts of `cat' is included in the set of contexts of `goldfish': 

\begin{center}
\begin{tabular}{c|ccc}
& aquarium & pet & fish\\
\hline
goldfish & 1 & 1 & 1 \\
cat & 0 & 1 & 1  
\end{tabular}
\end{center}

\noindent
By moving to a matrix setting, we are able to represent this correlation and get the correct entailment relation between the two words. In this case, the basis are pairs of the original basis elements.  Abbreviating them to their first letters,  the  matrix representations of `cat' and `goldfish' become:

\[
\begin{tabular}{c|ccc}
goldfish& a & p & f \\
\hline
a & 1 & 0 & 0 \\
p & 0 & 0 & 1 \\
f & 0 & 1 & 0
\end{tabular}
\qquad
\begin{tabular}{c|ccc}
cat& a & p & f \\
\hline
a & 0 & 0 & 0 \\
p & 0 & 1 & 0 \\
f & 0 & 0 & 1 
\end{tabular}
\]
It is easy to see that in this case the inclusion between the basis vectors, which now come in pairs, fails and as a result neither word entails the other. So we get a correct relationship. 

The above are not density matrices, we make them into such by using Equation \ref{eq:dm-qm}, as a result of which we obtain the following:
\[
\hat{\textmd{goldfish}} = \ov{\textmd{a}} \otimes\ov{\textmd{a}}  + (\ov{\textmd{p}}+\ov{\textmd{f}}) \otimes(\ov{\textmd{p}} + \ov{\textmd{f}})\qquad
\hat{\textmd{cat}} = (\ov{\textmd{p}}\otimes \ov{\textmd{p}})  + (\ov{\textmd{f}} \otimes \ov{\textmd{f}}) \]
 The explicit denotations of the  basis vectors are as follows:
\[
\ov{\textmd{a}} = (1,0,0) \qquad
\ov{\textmd{p}} = (0,1,0) \qquad
\ov{\textmd{f}} = (0,0,1)
\]
The resulting density matrices have the following tabular form:
\[
\begin{tabular}{c|ccc}
goldfish& a & p & f \\
\hline
a & 1 & 0 & 0 \\
p & 0 & 1 & 1 \\
f & 0 & 1 & 1 
\end{tabular}
\qquad
\begin{tabular}{c|ccc}
cat& a & p & f \\
\hline
a & 0 & 0 & 0 \\
p & 0 & 1 & 0 \\
f & 0 & 0 & 1 
\end{tabular}
\]
The lack of inclusion between these representations becomes apparent from Figure \ref{fig:graphs}, where it is shown that the subspaces spanned by the  basis vectors of the density matrices do not have an overlap.

\begin{figure}[t!]
\begin{center}
\includegraphics[scale=0.58]{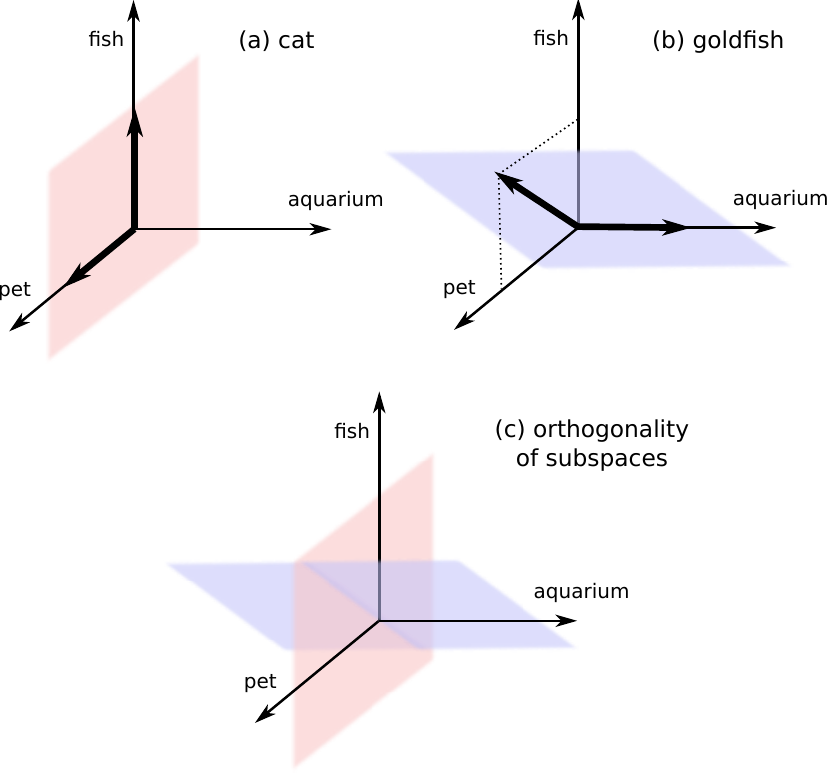}
\end{center}
\caption{Inclusion of subspaces in the `goldfish' example.}
\label{fig:graphs}
\end{figure}

\end{example}

Without taking correlations of the basis into account, DIH has been strengthened from another perspective and by the realization that  contexts should not be all treated equally. Various measures were introduced to weight the contexts based on their \emph{prominence},  for example by taking into account their rank  \cite{weeds2004,clarke2009,kotlerman2010}. From the machine learning side, classifiers have been trained  to learn the entailment relation at the word level \cite{baroni2012}. All of these  improvements are  applicable to the above density matrix  setting.

%

\section{Categorical Compositional Distributional Entailment}


The distributional co-occurrence hypothesis does  not naturally extend from the level of words to the level of sentences. One cannot mimic the basic  insights of the setting and say that sentences that have similar contexts have similar meanings, or that meaning of a sentence can be derived from the meanings of the words or sentences around it. The same fact holds about  the distributional  inclusion hypothesis and entailment, which does not naturally extend from words to phrases/sentences. One cannot say that a sentence $s_1$ entails a sentence $s_2$ when the contexts of $s_1$ are included in the contexts of $s_2$.  In the same lines, one cannot say that  two sentences entail each other if their meanings subsume each other. In this case, and similar to the case of co-occurrence distributions and similarity, entailment should  be computed compositionally.  

In this section, we define a compositional distributional notion of entailment based on the (vector and density matrix) representations of the words therein, the entailment relations between them,  and the grammatical structures of the sentences. This notion  is similar to the entailment-as-monotonicity notion of entailment in Natural Logic, which is based on an upward/downward monotonicity  relationship between the meanings of words  \cite{MacCartney2007}.  Whereas in Natural Logic grammatical structures of  sentences are treated on a case by case phrase-structure basis,  in our setting the strongly monoidal $F$ functor works in a modular and uniform fashion. 

Given a CCDS, in either of its   vectors or density matrices instantiations, we define a compositional notion of entailment, as follows:

\begin{definition}
{\rm Categorical compositional distributional entailment (CCDE)}.  For two strings $v_1 v_2 \cdots v_n$ and $ w_1 w_2 \cdots w_k$,  and $X$ either $KL$ or $N$, we have 
$v_1 v_2 \cdots v_n \vdash w_1 w_2 \cdots w_k$ whenever  
${R}_X(\semantics{v_1 \cdots v_n},\semantics{w_1 \cdots w_k}) \neq 0$.
\end{definition}

 We  show that this entailment can be made compositional for phrases and sentences that  have the same number of words and  the same grammatical structure and wherein the words entail each other point-wisely. We make this precise below. 

\begin{theorem}
\label{thm:main}
 For all $i, 1\leq i\leq n$ and $v_i, w_i$ 
 %
 words, we have
 \[
v_i \vdash w_i \quad \Rightarrow \quad v_1 v_2 \cdots v_n \vdash w_1 w_2 \cdots w_n
\]
whenever the  $v_1 v_2 \cdots v_n$ and $w_1 w_2 \cdots w_n$ have the same grammatical structure. 
\end{theorem}

\noindent
{\bf Proof.}
Consider the case of density matrices. By Eq.  \ref{densityentail}  and  CCDE, it suffices to show: 
\begin{equation}
\label{unfold}
\forall \hat{v}_i, \hat{w}_i \quad 
{R}_N(\hat{v}_i, \hat{w}_i) \neq 0 \ \implies \ 
{R}_N(\widehat{v_1 \cdots v_n},\widehat{w_1 \cdots w_n}) \neq 0
\end{equation}
By definition,   $\hat{R}(\hat{v}_i, \hat{w}_i) \neq 0$ is equivalent to the existence of $r_i \in \mathbb{R}$ and a positive operator $\hat{v'}_i$ such that $\hat{w_i} = r_i \hat{v_i} + \hat{v'_i}$.  Thus to prove the implication in  \ref{unfold} one can  equivalently  prove that there exist $r_i, q  \in \mathbb{R}$ and   positive operators $
\hat{v'}_i, \hat{\pi'}$ such that:
\[
\forall \hat{v}_i, \hat{w}_i  \quad  \hat{w_i} = r_i \hat{v_i} + \hat{v'_i} \ \implies \ 
\widehat{w_1 \cdots w_n} = q \cdot  \widehat{v_1 \cdots v_n} + \hat{\pi'}
\]
 According to the principle of lexical substitution with density matrices (Eq. \ref{denstlexsub})  we have:
 \[
 \widehat{v_1 \cdots v_n} := F(\alpha) (\hat{v_1} \otimes \cdots \otimes \hat{v_n}) + \hat{\pi'} \qquad
 \widehat{w_1 \cdots w_n}  :=   F(\beta) (\hat{w_1} \otimes \cdots \otimes \hat{w}_n) 
 \]
for $\alpha$ the grammatical structure of $\widehat{v_1 \cdots v_n}$ and $\beta$ the grammatical structure of $\widehat{w_1 \cdots w_n}$.  Thus what we want to prove  becomes  equivalent to the following:
 \[
 \forall \hat{v}_i, \hat{w}_i  \quad  \hat{w_i} = r_i \hat{v_i} + \hat{v'_i} \ \implies \ 
 F(\beta) (\hat{w_1} \otimes \cdots \otimes \hat{w}_n) = q  F(\alpha) (\hat{v_1} \otimes \cdots \otimes \hat{v_n}) + \hat{\pi'}
 \]
  In order to prove the above, we assume the antecedent and prove the consequence. That is, we assume  that  for all $\hat{v}_i$ and $\hat{w}_i$ there exist real numbers $r_i \in \mathbb{R}$ and  positive operators $\hat{v'}_i$, such that  $\hat{w_i} = r_i \hat{v_i} + \hat{v'_i}$ and prove the consequence. To prove the consequence,  we proceed as follow. Start from the assumption,  that  is for all $1\leq i \leq n$ we have
\[
\hat{v}_i \vdash \hat{w}_i 
\]   
This is equivalent to:
\[
\hat{v}_1 \vdash \hat{w}_1, \cdots, \hat{v}_n \vdash \hat{w}_n
\]
equivalent to:
 \[
 {R}_N(\hat{v}_1,  \hat{w}_1) \neq 0,   \cdots,  R_N(\hat{v}_n, \hat{w}_n) \neq 0
 \]
equivalent to:
\[
 \hat{w}_1 = r_1 \hat{v_i} + \hat{v'_1},  \cdots,     \hat{w}_n = r_n \hat{v}_n + \hat{v'}_n
\]
for $r_i$ and $\hat{v'_i}$ as defined previously. Using this, for  the tensor of  $\hat{w}_1$ to $\hat{w}_n$ we obtain:
\[
\hat{w_1} \otimes \cdots \otimes \hat{w}_n = ( r_1 \hat{v_i} + \hat{v'}_1) \otimes \cdots \otimes   ( r_n \hat{v}_n + \hat{v'}_n)
\]
which by bilinearity of tensor is equivalent to:
\[
r_1 \cdots r_n  (\hat{v_1} \otimes \cdots \otimes  \hat{v_n}) + \Pi
\]
where $\Pi$ is an expression of the following form:
\[
(r_1 \hat{v'}_1\otimes \hat{v}_2 \otimes \cdots \otimes \hat{v}_n) + 
(r_2 \hat{v}_1 \otimes \hat{v'}_2 \otimes \cdots \otimes \hat{v'}_n) + \cdots + 
(r_n \hat{v}_1 \otimes \hat{v}_2 \otimes \cdots \otimes \hat{v'}_n)
\]
Since the $\hat{v}_i$'s are density matrices (hence positive),  the $\hat{v'}_i$'s are positive operators, and summation and taking tensors preserves positivity, $\Pi$  is also a positive operator. Recall that $\widehat{v_1 \cdots v_n}$ and $\widehat{w_1 \cdots w_n}$ had the same grammatical structures, hence we have that $F(\alpha) = F(\beta)$. Denote this same structure with $f$.  We have:

\[
f(\hat{w_1} \otimes \cdots \otimes \hat{w}_n ) = f(r_1 \cdots r_n  (\hat{v_1} \otimes \cdots \otimes  \hat{v_n}) + \Pi)
\]
Since $f$ is a completely positive map, it is also linear, thus we have:
\[
 f(r_1 \cdots r_n  (\hat{v_1} \otimes \cdots \otimes  \hat{v_n}) + \Pi) = 
 r_1 \cdots r_n f (\hat{v_1} \otimes \cdots \otimes  \hat{v_n}) + f(\Pi)
\]
Since $f$ is completely positive $f(\Pi)$ is also positive. So we have shown:
\[ 
q  F(\alpha) (\hat{v_1} \otimes \cdots \otimes \hat{v_n}) + \hat{\pi'}
\]
for $q =  r_1 \cdots r_n$ and $ \hat{\pi'} := f(\Pi)$. 

The proof for  the case of vectors follows the same steps and it is simpler. In this case,  $\ov{v}_i \vdash \ov{w}_i$ is equivalent to $R_{KL}(\ov{v}_i, \ov{w}_i) \neq 0$, which is equivalent to the existence of $r_i \in \mathbb{R}$ and another vector $\ov{v'}_i$  such that $\ov{w}_i = r_i \ov{v}_i + \ov{v'}_i$.  Thus we drop the requirement about the existence of positive operators and  wherever it is used in the above, replace it with just a vector. In this case, the fact that $f$ is a linear map, i.e. a morphism in $ \textmd{FVect}_{\mathbb{R}} $ rather than ${\cal CPM}( \textmd{FVect}_{\mathbb{R}})$,  would suffice to get the required result. End of proof. $\square$

The above proposition means if  $w_1$ represents  $v_1$ and $w_2$ represents $v_2$ and so on  until $w_n$ and $v_n$, then the string $w_1 w_2 \cdots w_n$ represents the string $v_1 v_2 \cdots v_n$ compositionally,  from meanings of phrases/sentences. That is, the degree of representativeness of words -- either based on KL-divergence or von Neumann entropy -- extends to the degree of representativeness of phrases and sentences.   

\section{Working with Real Data}
 
The purpose of this section is twofold. First, we elaborate on the motivation of the `goldfish-cat' example (i.e. Example~\ref{eg:goldfish})  of  Section  \ref{sec:dih} and  present  five other cases of word  pairs and their co-occurrence counts from real data. Here our goal is to show that the correlation between the basis words, i.e. words corresponding to basis vectors, helps avoid unwanted entailments. Then, we present a  linguistic application of the proposed vector and density matrix models in a small-scale phrase/sentence entailment task based on data collected from a text corpus. 

\subsection{Correlation of Basis Words}

Our goal in this section is to ground  the  `goldfish-cat' example of Section \ref{sec:dih} in real data. That is, we  find pairs of words that would wrongly  entail each other in the vector view of the distributional hypothesis. Then, we find  basis words for these words in a way that these basis words correlate with each other. Finally,  we show that the corresponding density matrix representations of the words do not entail each other, or do so to a much lesser degree than the vector case.   We chose  the word pairs, the basis words,  and the co-occurrence counts from real data.  

In the first part of the experiment  we are  verifying two things. First is that whether  data  reflects the fact that whenever the first word in the pair occurred in the context of one of the basis words, was the other basis word  also present in the context window or not. Second, we want to show that the second word of the pair did occur with one of the basis words without the other one being around. The word pairs  and their correlated basis vectors are as follows:

\begin{center}
\begin{tabular}{c|c|c}
word pair & base 1 & base 2\\
\hline
\hline
(evidence, cigarette) & smoking & gun\\
(car, animal) & zebra & crossing\\
(bird, dancing) & night & owl\\
(goldfish, cat) & pet &fish\\
(BB, rifle) & toy & gun\\
(chlorine, fish) & swimming & pool
\end{tabular}
\end{center}

In order to ensure a correlation between the  basis words, we chose these in a way to form two-word non-compositional compound nouns, a list of some of which  is provided in \cite{Reddy11}. After choosing the basis words, we pick some target words. These word  pairs were chosen such that one of the words in the pair would be related to the meaning of the compound as a whole and the other word of the pair would be related to the meaning of only one of the words in the compound. For instance, in the first word pair, the word `evidence' is related to the meaning of the full compound, `smoking gun', whereas the word  ``cigarette' is related only to one of the nouns in the compound, in this case  to `smoking'.  Similarly, in the second pair, `car' is related to `zebra crossing' and `animal' just to `zebra'.  By means of example, what we aim to verify is that whenever `evidence' occurred in the same context with `smoking', `gun' was also around, but it was also the case that `cigarette' was present close to `smoking' without `gun' being around. Similarly for the other case, we want to verify that whenever `car' occurred in the same context with `zebra', the word `crossing' was around, but `animal' did occur with `zebra' without `crossing' being around. 

We collected co-occurrence counts for the pairs and the basis words. In all the example word pairs,  the vectors of the words have non-zero weights on both of the basis words, leading to inclusions of their contexts, indicating a wrong entailment relation between the two words of the pair. As an example, for the (evidence, cigarette) and (car, animal) pairs, the vector representations are as follows:
 
\begin{center}
\begin{tabular}{c|cc}
 & smoking &\qquad gun\\
\hline
evidence  & 1390 & 468 \\
cigarette & 4429 & 121  
\end{tabular} \hspace{2cm}
\begin{tabular}{c|cc}
 & zebra &\qquad  crossing\\
\hline
car  & 81 & 332 \\
animal & 389 & 44  
\end{tabular}
\end{center}
 
The matrix versions of these words were  indeed more indicative of the lack of an entailment relation within the pair.  In this case,   one of the words had a small number on its off diagonal entries and the other word had a larger number  there. For example, the matrix representations of the words of the  (evidence, cigarette) word pair  are as follows:
 \[
\begin{tabular}{c|cc}
 evidence& smoking &\qquad gun\\
\hline
smoking  & 1390 & 67  \\
gun & 67  & 468   
\end{tabular} \hspace{2cm}
\begin{tabular}{c|cc}
 cigarette& smoking &\qquad gun\\
\hline
smoking  & 4429 & 0  \\
gun & 0  & 121   
\end{tabular}
 \]
 The off diagonal counts are the counts for the basis pair (smoking,gun), i.e.  `evidence' was close to both  `gun' and `smoking'  for 67 times, whereas the cases where  `cigarette' was  close to both `smoking' and `gun' was 0.  This pattern is similar for the (car, animal) pair, but with less extreme non-zero off diagonal weights:  
 
  \[
\begin{tabular}{c|cc}
 car& zebra &\qquad crossing\\
\hline
zebra  & 81 & 11  \\
crossing & 11  & 332   
\end{tabular} \hspace{2cm}
\begin{tabular}{c|cc}
 animal& zebra &\qquad crossing\\
\hline
zebra  & 389 & 1  \\
crossing & 1  & 44   
\end{tabular}
 \]
 In this case, `car' was close to both `zebra' and `crossing' for  11 times, whereas this number for `animal' was only 1. We observed a  similar  pattern  for the other word pairs. In order to compare them, we normalised the off diagonal weights by dividing them by their sum and obtained a number between 0 and 1 for all the cases. These numbers are presented in the table below in decreasing order:
 
 \[
 \begin{tabular}{c|cc}
 word pair & off diagonal   &\qquad off diagonal \\
 &word 1 & \qquad word 2\\
 \hline
 \hline
 (evidence, cigarette) & 1.00 & \qquad 0.00 \\
 (car, animal)         & 0.91 & \qquad 0.09\\
 (bird, dancing)       & 0.85 & \qquad 0.15\\
 (goldfish, cat)       & 0.71 & \qquad 0.29\\
 (BB, rifle)           & 0.69 & \qquad 0.31\\
 (chlorine, fish)      & 0.56 & \qquad 0.44
 \end{tabular}
 \]
In all the cases,  the off diagonal ratios are more than 50\%  apart from each other, which indicates a less than 50\% overlap in their density matrix subspaces. Although real data is noisy, we do have a perfect separation: in the (evidence, cigarette) case, the off diagonal ratios are 100\% apart. This number decreases to about 90\% for (car,animal), to 85\% for (bird, dancing) and to 0.71\% for (goldfish, cat). The ratio of the  last two word pairs is lower than the rest, but still above 50\%. This is  because the compounds from which we derived the basis words for these pairs are not as non-compositional  as the other compounds. In other words, the word `pool' occurs many times on its own when it means `swimming pool' and the word `toy' is often dropped from the compound `toy gun' when talking about BB. 

Here,  we have only considered and provided data for modelling   correlations between pairs of basis. This can in theory be extended to   correlations between  $n$-tuples of basis, for any $n\geq 3$. In order to do so, one has to apply the ${\cal CPM}$ construction $n$ times, resulting in semantic categories $\underbrace{{\cal CPM}({\cal CPM}( \cdots ({\cal CPM})))}_n(\textmd{FVect}_{\mathbb{R}})$ and work with higher order density operators that embed in the extended spaces.
Providing real data for these general settings can be difficult due to sparsity problems, as one has to gather information about co-occurrences of $n+1$ words at the same time (the target word and the $n$-tuples of basis). A possible solution  to this problem is to take the limit of these co-occurrences as $n$ grows and only work until $n$'s that allow for gathering  reasonable quantities of co-occurrence data. Choosing the number to which $n$  tends to is related to the existence of $n$-word non-compositional compounds in language. In principle, this number can grow arbitrarily large,  as for any $n$-word such compound, one is able to create a larger one with $n+1$ words. In practice,  however, text corpora contain data for $n$'s that are small (usually not greater than 2 or 3).

\subsection{Toy Entailment Application}

\paragraph{\bf Dataset.}
In order to create our dataset, we first randomly selected 300 verbs from the most frequent 5000 words in the British National Corpus,\footnote{BNC is a 100 million word collection (around 2 million sentences) of samples of written and spoken language from a wide range of sources, available at {\tt http://www. natcorp.ox.ac.uk/}} and randomly picked either a hyponym or a hyponym from WordNet, provided that these also  occurred more than 500 times in the BNC. Next, each entailing verb was paired with one of its  subject or object nouns, which had again occurred more than 500 times. The corresponding entailed verb was paired with an appropriate hypernym of this noun chosen from the set described above. Recall that one has the following entailment between the hyponyms and the  hypernyms:   
\[
\mbox{\it hyponym} \  \vdash  \ \mbox{\it hypernym}
\]
This procedure created 300 phrase/sentence entailments of the form 
\begin{center}
\begin{tabular}{ccc}
entry 1 & $\vdash$ & entry 2 \\
\hline
{\em subject$_1$ verb$_1$} &$\vdash$& {\em subject$_2$ verb$_2$}\\

  {\em verb$_1$ object$_1$} &$\vdash$&  {\em verb$_2$ object$_2$}. 
 \end{tabular}
 \end{center}

 Many of these 300 pairs did not reflect an easily recognisable entailment. As our goal was to collect human judgements for the degrees of entailments, we had to have pairs in which the entailment or lack thereof was obvious for humans. Thus,  from these 300 entries, we selected 23  pairs to reflect three ranges of entailment degrees, classified as follows:
 
 \begin{enumerate}
 \item Both the subjects (or objects) and the verbs   entail each other respectively, that is:
  \[
 subject_1 \vdash subject_2 \qquad \text{and} \qquad verb_1 \vdash verb_2
  \]
    \[
 object_1 \vdash object_2 \qquad \text{and} \qquad verb_1 \vdash verb_2
  \]
 \item  Either the subjects (or objects)  entail each other or the verbs do, that is
 \[
 subject_1 \vdash subject_2 \qquad \text{or} \qquad verb_1 \vdash verb_2
 \]
  \[
 object_1 \vdash object_2 \qquad \text{or} \qquad verb_1 \vdash verb_2
 \]
 
 \item  Neither the subjects (or objects)  nor the verbs entail each other (or at least they did not do so  in a clear way), that is
 \[
 subject_1 \nvdash subject_2 \qquad \text{and} \qquad verb_1 \nvdash verb_2
 \]
 \[
 object_1 \nvdash object_2 \qquad \text{and} \qquad verb_1 \nvdash verb_2
 \]
 
 \end{enumerate}
 
 Whereas the pairs created by the above procedure cover entailments between short two-word phrases and sentences, we were also interested in providing results for full transitive sentences. In order to do that, we used the 23 pairs to form subject-verb-object entailments by following the procedure below:

 \begin{itemize}
 \item
   pairing the subject of an intransitive sentence and its hypernym with a verb phrase and its hypernym, for example    `people' in `people strike'  was paired with `group'  in `group attacks' and `clarify rule' was paired with `explain process',  
  \item 
   pairing the object of a verb phrase and its hypernym with an intransitive sentence and its hypernym, for example `task' in  `arrange task' was paired with `work' in `organise work' and `notice advertise' was paired with `sign announce'. 
  \end{itemize}
  
  Similar to the intransitive sentence and verb phrase case, we  went through the resulting sentences and chose 12 of them that had either easily recognisable entailments for humans or were obviously not entailing each other, again relative to the human eye. These reflected three ranges of  entailment degrees classified as follows:
 
\begin{enumerate}
\item Both the subjects (or objects) and the verb phrases (or the intransitive sentences) entailed  each other, that is:
  \begin{eqnarray*}
   subject_1 \vdash subject_2 &\qquad \text{and} \qquad & verb \ \ phrase_1 \vdash verb \ \  phrase_2\\
   object_1 \vdash object_2 &\qquad \text{and} \qquad & intr. \ \ sentence_1 \vdash intr. \ \ sentence_2\\
  \end{eqnarray*}
  \item  Either the subjects (or objects) or the verb phrases (or the intransitive sentences) entailed  each other, that is:
  \begin{eqnarray*}
   subject_1 \vdash subject_2 &\qquad \text{or} \qquad & verb \ \ phrase_1 \vdash verb \ \ phrase_2\\
   object_1 \vdash object_2 &\qquad \text{or} \qquad & intr. \ \ sentence_1 \vdash intr. \ \ sentence_2\\
  \end{eqnarray*}
  
    \item  Neither the subjects (or objects) nor the verb phrases (or the intransitive sentences) entailed  each other, that is:
  \begin{eqnarray*}
   subject_1 \nvdash subject_2 &\qquad \text{and} \qquad & verb \ \ phrase_1 \nvdash verb \ \ phrase_2\\
   object_1 \nvdash object_2 &\qquad \text{and} \qquad & intr. \ \ sentence_1 \nvdash intr. \ \ sentence_2\\
  \end{eqnarray*}
  \end{enumerate}

The degree of entailment between the produced phrases and sentences were evaluated by 16 annotators. These were either logic or computational linguistics professionals. They provided their scores in a scale from 1 (no entailment) to 7 (full entailment). The 1-7 scale was chosen following  common practice in the empirical computational linguistics literature, for example see  \cite{lapata2010}. Each entailment was scored by the average across all annotators. The human judgements agreed with the three classes of entailments, described above. That is, we had three clear bands of judgements:

\begin{enumerate}

\item The entries in which both subjects/objects  and verbs/verb phrases/intransitive sentences entailed each other, got an average annotation  above 4. For example we had:

\small
\begin{center}
\begin{tabular}{l|rcl|cc}
Entry & entry 1  & $\vdash$&  entry 2  & Avg. judgement\\
\hline\hline

\multirow{2}{*}{intr. sentence} 

&people strike   & $\vdash$&  group attacks  & 4.313\\

&notice advertises   & $\vdash$& sign announces  & 5.375\\

\hline
\multirow{2}{*}{verb phrase}

&clarify rule & $\vdash$&  explain process  &5.000\\

&recommend development   & $\vdash$& suggest improvement & 5.375\\

\hline
\multirow{2}{*}{trans. sentence}

&people clarify rule  & $\vdash$&   group explain process & 5.000\\


& office arrange task & $\vdash$& staff organize work & 5.500\\

\end{tabular}
\end{center}
\normalsize

\item The entries  in which  either only subjects/objects entailed each other or only verbs/verb phrases/intransitive sentences did, got an average annotation between 1 and 4. For example:

\small
\begin{center}
\begin{tabular}{l|rcl|cc}
Entry & entry 1  & $\vdash$& entry 2  & Avg. judg.\\
\hline\hline

\multirow{2}{*}{intr. sentence}

&corporation appoints & $\vdash$&firm founds & 3.313\\

&boy recognizes & $\vdash$& man remembers& 2.938\\

\hline
verb phrase
& confidence restores & $\vdash$& friendship renews& 2.625\\

\hline
\multirow{2}{*}{trans. sentence}

& corporation appoint people & $\vdash$& firm found group & 2.937\\
& people read letter & $\vdash$& corporation anticipate document & 2.062

\end{tabular}
\end{center}
\normalsize

In the first case,   `corporation' clearly entails `firm', but  the entailment relationship between `appoints' and `founds' is unclear.  In the second case, clearly `boy' entails `man', but it is not so obvious if `recognise' entails `remember'.   In the third case, again `restores' clearly entails `renews', but the relationship between `confidence' and `friendship' is less evident. In the fourth case, `corporation' clearly entails `firm', but  the relationship between `appoint people' and `found group' is not very obvious. 

\item  The entries which were non-entailing, i.e. it was not clear if we had an entailment relationship between the subjects/objects and it was not clear if we had an entailment relationship between the verbs/verb phrases/intransitive sentences, got an average annotation below 2. For example:

\small
\begin{center}
\begin{tabular}{l|rcl|cc}
Entry & entry 1  & $\vdash$& entry 2  & Avg. judgement\\
\hline\hline

\multirow{2}{*}{intr. sentence}

&editor threatens & $\vdash$& application predicts, & 1.125\\
& progress reduces & $\vdash$&development replaces& 1.225\\

\hline
verb phrase 

& confirm number & $\vdash$& approve performance & 1.813\\

\hline
\multirow{2}{*}{trans. sentence}

& editor threatens man & $\vdash$& application predicts number &  1.125\\
&man recall time & $\vdash$&  firm cancel term & 1.625
\end{tabular}
\end{center}
\normalsize

Consider for example the fourth entry:  it is clear that neither `editor' entails `application', nor `threatens man' entails `predicts number'. Similarly, in the third entry, `confirm' does not entail `approve' and `number' does not entail `performance'. Also similarly in the first case, it is clear that  `editor' does not entail `application' and neither does `threatens' entail `predicts'. 
\end{enumerate}

\paragraph{\bf Basic vector space.} The distributional space where the vectors of the words live is a 300-dimensional space produced by non-negative matrix factorization (NMF). The original vectors were 2,000-dimensional vectors weighted by local mutual information (LMI), for which the contexts counts had been collected from a 5-word window around each target word. The vectors were trained on the concatenation of ukWaC and Wackypedia corpora.\footnote{Around 132 million sentences of English text, available at {\tt http://wacky.sslmit. unibo.it/doku.php?id=corpora}/}


\paragraph{\bf Entailment via KL-divergence in $\textmd{FVect}_{\mathbb{R}}$.} 
For degrees of entailment obtained via KL-divergence,  we work on the instantiation of CCDS to $\textmd{FVect}_{\mathbb{R}}$ for the three types of phrases/sentences in our dataset: 

\begin{enumerate}
\item verb phrases,  which we will refer to by  ``\emph{verb noun}'',
\item  intransitive sentences, which we will refer to by  ``\emph{noun verb}", 
\item   transitive sentences, which we will refer to by  ``\emph{noun verb noun$'$}". 
\end{enumerate}
The vector representations of these are  obtained by applying Equation \ref{eq:lexsub}, which result in the following expressions:
\begin{gather}
\label{eq:verbnoun-KL}
\ov{\mbox{verb noun}} := F(\alpha)(\ov{v} \otimes \ov{n}) = (1_S \otimes \epsilon_N) (\ov{v} \otimes \ov{n}) \\
\ov{\mbox{noun verb}} := F(\alpha)(\ov{n} \otimes \ov{v}) = (\epsilon_N \otimes 1_S) (\ov{n} \otimes \ov{v}) \\
\ov{\mbox{noun verb noun}'} :=  F(\alpha)(\ov{n} \otimes \ov{v} \otimes \ov{n'}) = (\epsilon_N \otimes 1_S \otimes \epsilon_N) (\ov{n} \otimes \ov{v} \otimes \ov{n'})
\end{gather}
The first two of the above items simplify to the matrix multiplications between the matrix of the verb and the vector of the noun, as follows,  for $\ov{n}^{\text T}$ the transpose of the vector of the noun:
\begin{gather}
\label{eq:matrixmult}
\ov{v} \times \ov{n}\\
\ov{n}^{\text T} \times \ov{v}
\end{gather}
 The vector representation of a ``\emph{noun verb noun$'$}'" sentence simplifies to the tensor contraction between the cube of the verb and the vector of noun$'$, and then the matrix multiplication between the matrix of the result and the vector of the noun, as follows:
 \begin{equation}
 \label{eq:cubecontr}
  \ov{n}^{\text T} \times \ov{v} \times \ov{n'}
  \end{equation}
 For details of these computations, we refer the reader to our  previous work \cite{CoeckeSadrClark2010,grefenstette2011,KartSadrPul-COLING-2013}, where these and other forms of sentences have been worked out for a variety of different nouns and verbs, as well as adjectives (for sentences with adjectival modifiers).  

\paragraph{\bf Matrices and Cubes of Verbs.}
Vectors of nouns $\ov{n}$ are created using the usual distributional method.  For producing the verb matrices for verbs taking a single argument (either at the subject or the object position), we work with a  variation of the method suggested in \cite{grefenstette2011}, referred to by \emph{relational}. Specifically, we define the verb matrix as follows:
\begin{equation}
\label{eq:intrans-verb}
\ov{v}_{matrix} =  \ov{v} \odot \sum_i ( \ov{n}_i \otimes \ov{n}_i) \\
\end{equation}

In the above, $\ov{n_i}$ enumerates {\em all} the nouns that the verb has modified across the corpus in various phrases and sentences. $\ov{v}$ is the distributional vector of the verb, built in the same way as the noun vectors.  
The original relational method computed the matrix of the verb by encoding in it  the information about the noun arguments of the verb across the corpus, the same as we do above. The above formulation enriches this encoding,   via the use of  the point-wise multiplication operation $\odot$,  by also taking into account the distributional vector of the verb $\ov{v}$, hence  encoding directly information about the distributions of the  verb. Substituting this in  the  matrix multiplication of the expression in Eq. \ref{eq:matrixmult} and simplifying it,  provides us with the following  vector  representation for ``verb-noun" and ``noun-verb" expressions:

\begin{equation}
\label{eq:phrase}
 \ov{\mbox{noun verb}}  =  \ov{\mbox{verb noun}} =  \ov{v} \odot \sum_i \langle \ov{n} \mid \ov{n}_i \rangle \ov{n}_i
\end{equation}

Roughly speaking, the above says that the vector meaning of any such phrase/sentence represents  the contextual properties of the verb of the phrase together with the common contextual properties of the nouns of the phrase/sentence and the nouns that the verb has modified across the corpus. 

In order to represent the meaning of transitive verbs, the matrices of Equation \ref{eq:intrans-verb} are embedded  into cubes $\textbf{C}_{ijk}$ by  copying either their $i$'th  or their $j$'th dimension into the extra $k$'th dimension of the cube.  Thus obtaining the following two cubes:

\[
\textbf{C}_{iij}  \qquad \text{and} \qquad \textbf{C}_{ijj} 
\]

This operation is formally referred to as a Frobenius algebra copying operation and is extensively  discussed and applied in previous work, e.g.  \cite{KartSadrPul-COLING-2013,kartsaklisphd,piedeleu2015,KartSadr-EMNLP-2013,milajevs2014,Kartsaklis16}.  The first embedding (providing us with the cube $\textbf{C}_{iij}$)  is referred to as \emph{copy subject} and the second one (providing us with the cube $\textbf{C}_{ijj}$)  as \emph{copy object}. The sentence vectors produced by the two methods when we substitute such cubes in Equation \ref{eq:cubecontr} take the following form:

\begin{gather}
   \text{Copy Subject:~} (\ov{v}_{matrix} \times \ov{n}_{object}) \odot \ov{n}_{subject} \\
   \text{Copy Object:~} (\ov{v}_{matrix} \times \ov{n}_{subject}) \odot \ov{n}_{object}
\end{gather}

\noindent
where $\ov{n}_{subject}$, $\ov{n}_{object}$ are the distributional vectors for the subject and the object of the transitive sentence, and $\ov{v}_{matrix}$ the matrix of the verb, in our case created as in Eq. \ref{eq:intrans-verb}. 

Since each one of the above embeddings puts emphasis on a different argument of the transitive verb, it is reasonable for one to represent the meaning of the transitive sentence by further combining both of them into a single representation, for example as below:

\begin{equation}
   \ov{v}_{CopySubject} + \ov{v}_{CopyObject}~~~~~~~
   \ov{v}_{CopySubject} \odot \ov{v}_{CopyObject}
   \label{eq:frob-merge}
\end{equation}

\paragraph{\bf Entailment via relative entropy in ${\cal CPM}(\textmd{FVect}_{\mathbb{R}})$.}  
In the case of degrees of entailment using relative entropy, we work with  the instantiation of CCDS to ${\cal CPM}(\textmd{FVect}_{\mathbb{R}})$, where Equation \ref{eq:lexsub} results in a density matrix, computed as follows for a ``verb noun"  phrase,  a ``noun verb"  and  a ``noun verb noun$'$" sentence, respectively:
\begin{eqnarray}
\label{eq:verbnoun}
\hat{\mbox{verb noun}} &:=& F(\alpha)(\hat{v} \otimes \hat{n}) = (1_S \otimes  \epsilon_N) (\hat{v} \otimes \hat{n}) \\
\hat{\mbox{noun verb}} &:=& F(\alpha)(\hat{n} \otimes \hat{v}) = (\epsilon_N  \otimes 1_S) (\hat{n} \otimes \hat{v})\\
\hat{\mbox{noun verb noun}'} &:=& F(\alpha)(\hat{n} \otimes \hat{v} \otimes \hat{n'}) = (\epsilon_N  \otimes 1_S \otimes \epsilon_N) (\hat{n} \otimes \hat{v} \otimes \hat{n'})
\end{eqnarray}
where $\hat{v}, \hat{n}$ and $\hat{n'}$ are the density matrices of the verb and the nouns, respectively, and $\otimes$  is the tensor product in ${\cal CPM}(\textmd{FVect}_{\mathbb{R}})$. These simplify to the following formulae:
\begin{gather}
\label{eq:trace1}
 \text{Tr}_N (\hat{v} \circ (\hat{n}\otimes 1_S)) \\
\label{eq:trac2}
 \text{Tr}_N ((\hat{n}\otimes 1_S) \circ \hat{v})\\
\label{eq:trac3}
  \text{Tr}_{N,N}(\hat{v}\circ (\hat{n}\otimes 1_S \otimes \hat{n'} ) )
\end{gather}

\noindent
For details of the computations and examples with different nouns and verbs and sentence forms,  see Piedeleu et al.  \cite{piedeleu2015} and Kartsaklis \cite{kartsaklisphd}. 


The above formulae and equations of density matrices for words, phrases and sentences are theoretical.  In what follows, we implement two concrete ways of creating density matrices, one directly based on the correlations between the bases and another by algebraically operating on the vectors. 

\paragraph{\bf Density matrices by direct correlation.}
We describe a generic process for creating density matrices based on correlations between the basis vectors, similar to those demonstrated in the `goldfish' example of Section \ref{sec:dih} and depicted graphically in Figure \ref{fig:graphs}. Co-occurrence counts are collected for a target word $w$ and every pair of words $(w_i,w_j)$ (not necessarily in sequence) that occur in the same context with $w_t$. By using these statistics  and treating the pairs of words as a single basis, one can build an upper- or lower-triangular matrix, let us denote it by  $\textbf{M}$. Since the statistics were correlated regardless of the order of the words $w_i$ and $w_j$, we  can expand $\textbf{M}$ to a symmetric matrix. This is a routine procedure and is done  by copying  the upper or the lower  triangle into the other half of the matrix. Formally speaking, we have: 
\[
\textbf{M}_{ij} = \textbf{M}_{ji}
\]

In order for the matrices to be  density matrices, they have to be positive semi-definite. This can be enforced in different ways, one of which is by turning $\textbf{M}$ to a row \textit{diagonally dominant} matrix. This is a matrix  for every $i$-th row of which we have:

\begin{equation*}
\textbf{M}_{ii} \geq \sum_{i\neq j} |\textbf{M}_{ij}|
\end{equation*}

\noindent
That is,  in all of the rows of this matrix,   the magnitude of the diagonal entry is greater than or equal to the sum of the magnitudes of the non-diagonal entries. In our case, since the non-diagonal entries are counts, they are positive, and thus the entries and their magnitudes are equal.  We then normalise this matrix  by  its trace and obtain a   density matrix.


\paragraph{\bf Density matrices from distributional vectors.} In contrast with the previous section, the construction we present here follows directly the quantum-mechanical intuition expressed in Equation \ref{eq:dm-qm} that a density matrix is a probability distribution over a set of vectors. For a target word $w$, we define this set $\{\ov{c_i}\}_i$ to consist of vectorial representations of the various contexts in which $w$ occurs: for example, $\ov{c_i}$ can be the average of the distributional vectors for all other words in the same sentence with $w$. 
In symbols, the density matrix corresponding to a word $w$ is defined as follows:
\begin{equation}
  \hat{w} = \sum_i p_i\  \ov{c_i} \otimes \ov{c_i}
  \end{equation}
where $i$ iterates through all contexts of word $w$.

\paragraph{\bf Density matrices for transitive verbs.} The Frobenius embeddings (briefly discussed for the case of standard verb matrices above) can be also applied on the density matrix formulation, producing the following representations for transitive sentences:

\begin{gather}
   \text{Copy Subject:~} \hat{subj} \odot \text{Tr}_{N,N}(\hat{v}\circ (1_N \otimes \hat{obj})) \\ 
   \text{Copy Object:~} \text{Tr}_{N,N}(\hat{v}\circ (\hat{subj} \otimes 1_N)) \odot \hat{obj}
\end{gather}

\noindent where $\hat{v}$, $\hat{subj}$, and $\hat{obj}$ are density matrices created using one of the two methods (by direct correlation or from distributional vectors) presented above. Note that merging the two representations into one as in Equation \ref{eq:frob-merge} is also possible, since both element-wise addition and element-wise multiplication of two density matrices preserve the underlying structure. 

\paragraph{\bf From word to phrase and sentence density matrices.}
Substituting a word density matrix in Equations \ref{eq:trace1} to \ref{eq:trac3} and simplifying, results in the following density matrix representations for each phrase/sentence:
\begin{eqnarray}
\label{eq:simple-intans}
  \hat{\mbox{noun verb}} =   \hat{\mbox{verb noun}}  &=& \hat{v}^{\text T} \times \hat{n} \times  \hat{v}\\
  \hat{\mbox{noun verb noun}'} &=& \hat{v}^T \times (\hat{n}\otimes \hat{n'}) \times \hat{v}
\end{eqnarray}

Again, the formulation is the same for a ``\emph{verb noun}" and a ``\emph{noun verb}" phrase. In simple terms, the above result in  density matrices that take into account the contextual properties of the verb, the contextual properties of the nouns of the phrase/sentence, and those of the  nouns that the verb has modified across the corpus, with the added value that these properties now reflect correlations between the various contexts through the use of density matrices.

\paragraph{\bf Entailment for simple vector composition.} 
Finally, as a comparison, we also work with degrees of entailment obtained by computing KL-divergence on  a simple compositional model achieved via element-wise addition and element-wise multiplication of the vectors of the words in the phrase: 
\begin{eqnarray*}
 \ov{\mbox{noun verb}}_{+}  =   \ov{\mbox{verb noun}}_{+} = \ov{v} + \ov{n} &\qquad&    \ov{\mbox{noun verb}}_{\odot} =  \ov{\mbox{verb noun}}_{\odot} = \ov{v} \odot \ov{n}\\
  \ov{\mbox{noun verb noun}'}_{+} = \ov{n} + \ov{v} + \ov{n'} &\qquad&    \ov{\mbox{noun verb noun}'}_{\odot} = \ov{n} \odot \ov{v} \odot \ov{n'}
\end{eqnarray*}
where $\ov{v}$ and $\ov{n}, \ov{n'}$ denote the distributional vectors of the verb and  the nouns, respectively. 

The experiment proceeds as follows: We firstly produce phrase/sentence vectors and density matrices by composing the vectors or the density matrices of the individual words in each phrase, and then we compute an entailment value for each pair of phrases; in the case of vectors, this value is given by the representativeness on the KL-divergence between the phrase vectors, while for the density matrix case it is the representativeness on the von Neumann entropy between the density matrices of the phrases/sentences. The performance of each model is expressed as the Spearman's correlation of the model predictions with the human judgements. 

The results for the verb phrase/intransitive sentence entailment are presented in Table \ref{tbl:results1}. A non-compositional baseline is also included: we computed $R_{KL}$ for the lexical vectors of the heads of the sentences, that is their verbs. The upper bound is the inter-annotator agreement.

{\small
\begin{table}[h]
\begin{center}\begin{tabular}{|l|c|ccc|}
\hline
{\bf Model} & {\bf $\rho$} & {\bf Inf} & {\bf F1} &  {\bf Acc} \\
\hline \hline
Baseline  (vector of verb) & 0.24 & 0.37 & 0.57 & 0.74 \\
\hline
Categorical &&&&\\
\quad $R_{KL}$ (vectors) &{\bf 0.66} & 0.56 & 0.74 & 0.78\\
\quad $R_N$ (density matrices by direct correlation) & 0.42 & {\bf 0.67} & {\bf 0.80} & {\bf 0.87}\\
\quad $R_N$ (density matrices from vectors) & 0.48 & 0.60 & 0.76 & 0.78 \\
\hline
Simple&&&&\\
\quad $R_{KL}^+$ (e.w. addition) & 0.52 & 0.52 & 0.71 & 0.78 \\
\quad $R_{KL}^{\odot}$ (e.w. multiplication) & 0.41 & 0.32 & 0.64 & 0.61 \\
\hline
Upper bound & 0.66 & & & \\
\hline
\end{tabular}\end{center}
\caption{Results for the verb phrase/intransitive sentence entailment experiment.}
\label{tbl:results1}
\end{table}
}

We also present informedness, F1-score and accuracy for a binarised variation of the task, in which a phrase/sentence pair is classified as ``entailment'' or ``non-entailment'' depending on whether its average human score was above or below the mean of the annotation range. Informedness is an information-theoretic measure that takes into account the true negatives count (something that is not the case for F1-score, for example) and thus it is more appropriate for small and relatively balanced datasets such as ours. The numbers we present for the binary task are based on selecting an appropriate threshold for each model, above of which entailment scores are classified as positive. This threshold was selected in order to optimize informedness. 

The results show that all the compositional models (for both vectors and density matrices) outperformed the non-compositional baseline. In the correlation task, the categorical vector model  $R_{KL}$  was better, achieving a score that matches the inter-annotator agreement; in the classification task, the categorical density matrix models  $R_N$  are ahead in every measure. From the two density models we implemented, the one based on distributional vectors (Equation \ref{eq:dm-qm}) has a better degree of correlation with human judgements, but the one that directly reflects basis correlation presents the best binary performance, with accuracy 0.87 and informedness 0.67.

A snapshot of the  results including the highest and lowest pairs according to human judgements are shown in Table \ref{tbl:snapshot}. We see that although each model returns values in a slightly different range, all of them follow to some extent the general pattern of human annotations. From all three models, the predictions of the model based on element-wise multiplication of vectors are quite marginal. The categorical models and addition of vectors return more balanced results, without avoiding small mistakes.

\begin{table*}[h!]
\begin{center}
\begin{scriptsize}
\begin{tabular}{|l||c||c|c||c|c|}
\hline
{\bf Entailment} & {\bf Humans} & \multicolumn{2}{c||}{\bf Categorical} &\multicolumn{2}{c|}{\bf Simple} \\
&&$R_{KL} (0.12)$ & ${R_{N} (0.17)}$ & $R_{KL}^+$ (0.13) & $R_{KL}^\odot$ (0.08) \\
\hline
 arrange task $\vdash$ organize work              & 5.50 (0.785) - T &0.164 - T & 0.371 - T & 0.192 - T & 0.142 - T \\
recommend development $\vdash$ suggest improvement & 5.38 (0.768) - T & 0.146 - T  & 0.250 - T & 0.182 - T & 0.084 - T \\
advertise notice $\vdash$ announce sign            & 5.38 (0.768) - T & 0.114 - F & 0.187 - T & 0.100 - F & 0.090 - T\\
\hline
confirm number $\vdash$ approve performance       & 1.81 (0.258) - F &    0.111 - F & 0.140 - F & 0.087 - F & 0.084 - T \\
recall time $\vdash$ cancel term               & 1.63 (0.232) - F &0.070 - F    & 0.169 - F & 0.126 - F & 0.072 - F \\
editor threathen $\vdash$ application predict    & 1.13 (0.161) - F &0.082 - F   &0.184 - T & 0.092 - F & 0.080 - F \\
\hline
\end{tabular}
\end{scriptsize}
\end{center}

\caption{A snapshot of the phrase entailment experiment. The human judgements are between 1 and 7,  with their values normalised between 0 and 1 in brackets. The model predictions are between 0 and 1. T and F indicate classification of each phrase pair as entailment or non-entailment according to each model. The numbers that appear in brackets at the headers are the classification thresholds optimizing informedness for the various models. $R_N$ refers to the density matrix model based on word vectors.}
\label{tbl:snapshot}
\end{table*}

Table \ref{tbl:results-svo} presents the results for a transitive entailment experiment, based on the 12 subject-verb-object entailments created as described earlier in this section. 
We have not a similar table to Table 2 for transitive cases, since we have many more models for the transitive case and  most of these models acquired the same score (0.83/0.92) due to the small size of the dataset.
For the categorical compositional models we apply the Frobenius embeddings as described earlier, and combinations of these. For the density matrix formulation we use density matrices created from vectors, since this method showed better correlation with human judgements for the intransitive sentence entailment task. The results follow a pattern very similar to that of the intransitive sentence/verb phrase entailment experiment: For the correlation task, the highest performance comes from a  categorical model using standard matrices and vectors, specifically the Frobenius additive model (copy subject + copy model); this model presents a correlation 0.72, very close to the inter-annotator agreement (0.75). However, the highest performance in the classification task comes once more from density matrix models, exactly as in the previous experiment. On the other hand, this time some of the other models scored lower than the non-compositional baseline, possibly demonstrating an amount of correlation between sentence length and effectiveness of the model.


{\small
\begin{table}[h]
\begin{center}\begin{tabular}{|l|c|ccc|}
\hline
{\bf Model} & {\bf $\rho$} & {\bf Inf} & {\bf F1} &  {\bf Acc} \\
\hline \hline
Baseline  (vector of verb) & 0.40 & 0.62 & 0.75  & 0.83  \\
\hline
Categorical &&&&\\
\quad $R_{KL}$ Copy-subject & 0.43 & 0.12 & 0.33 & 0.67 \\
\quad $R_{KL}$ Copy-object  & 0.42 & 0.62 & 0.75 & 0.83 \\
\quad $R_{KL}$ Copy-subject $+$ Copy-object  & {\bf 0.72} & 0.62 & 0.75 & 0.83 \\
\quad $R_{KL}$ Copy-subject $\odot$ Copy-object  & 0.70 & 0.62 & 0.75 & 0.83 \\
\hline
\quad $R_N$ (density matrices from vectors, Copy Subject) & 0.38 & {\bf 0.75} & {\bf 0.86} & {\bf 0.92}  \\
\quad $R_N$ (density matrices from vectors, Copy Object) & 0.26 & 0.62 & 0.75 & 0.83  \\
\quad $R_N$ (density matrices from vectors, Copy Subject $+$ Copy Object) & 0.34 & {\bf 0.75} & {\bf 0.86} & {\bf 0.92}  \\
\quad $R_N$ (density matrices from vectors, Copy Subject $\odot$ Copy Object) & 0.06 & 0.62 & 0.75 & 0.83  \\
\hline
Simple&&&&\\
\quad $R_{KL}^+$ (e.w. addition)       & 0.68  & 0.62 & 0.75  & 0.83  \\
\quad $R_{KL}^{\odot}$ (e.w. multiplication) & 0.14  & 0.38 & 0.57 & 0.75 \\
\hline
Upper bound & 0.75 & & & \\
\hline
\end{tabular}\end{center}
\caption{Results for the transitive sentence entailment experiment.}
\label{tbl:results-svo}
\end{table}
}

\section{Conclusion and Future Directions}

We reviewed the categorical compositional distributional semantic (CCDS) model, which extends the distributional hypothesis from words to strings of words. We showed that the model can also extend the distributional inclusion hypothesis (DIH) from words to phrases and sentences. In this case, one is able to derive entailment results over strings of words, from the entailments that hold between their constituent words. We recalled how the vector-based CCDS, which normally works with the category of finite-dimensional vector spaces and linear maps $\textmd{FbVect}_{\mathbb{R}}$, can be extended to include density matrices and completely positive maps, by moving to the category ${\cal CPM}(\textmd{FVect}_{\mathbb{R}})$. We  reviewed the existing notion of KL-divergence and its application to word level entailment on vector representations of words. We then argued for and showed that  moving from vectors to density matrices strengthens the DIH. 

As contributions,  on the theoretical side we proved  that strings of words whose words point-wisely entail each other and where the strings  have the same grammatical structure, admit a compositional notion of entailment. This is an extension of the result of the conference version of this paper \cite{ISAIM16}, where a similar proof was presented for phrases and sentences which had grammatical structures that only consisted of epsilon-maps and identities. The previous result naturally excluded the cases where Frobenius or bialgberas are needed, e.g. for relative pronouns, as shown in \cite{SadrClarkCoecke1,SadrClarkCoecke2}, for coordination and intonation as shown in \cite{Kartsaklis16,KartSadr15}, and for quantification, as shown in \cite{HedgesSadr16}. The general version of the  theorem proved in this paper is applicable to all these cases. 

On the experimental side, we presented two small scale tasks,  both performed on real data. First, we presented evidence that density matrices do indeed give rise to a richer notion of entailment at the word level. This evidence consisted of pairs of words   whose vector representations, built from real data,  indicated a false entailment between the words, but where  their density matrices, also built from real data,  corrected the problem. 
Second, we built  vector and density matrix representations for short phrase/sentences, computed the  KL divergence and entropy between pairs of them  and  applied the results  to a phrase/sentence entailment task. Our dataset consisted of  pairs of intransitive sentences, object-verb phrases, and transitive sentences. The theoretical argument of the paper favours categorical composition over simple element-wise operators between vectors, and our results were supportive of this. The density matrices formulation  worked better on the classification task. For correlation between the degrees of entailment as predicted by the model and as judged by humans, the composition over standard matrices and vectors performed better. For the intransitive/verb-phrase entailment task, the concrete CCDS instantiations on vectors and density matrices  performed clearly above the baseline, while for the transitive sentence entailment task, some models scored lower than the baseline due to the increased complexity and the greater sentence lengths.
%
%
A large scale experiment to confirm these predictions constitutes work in progress. 

Theorem  \ref{thm:main} showed a relationship between the CCDS meanings of words (represented by vectors or density matrices), the corresponding word-level entailments thereof, and the grammatical structures of sentences. The proven relationship is, however, restrictive. It only holds for sentences that have the same grammatical structure. Studying this restriction and extending the theorem to a general form is work in progress. We aim  to prove a similar relationship between sentences that do not necessarily have the same grammatical structure, but that a possibly  weaker relationship holds between their grammatical structures. Note however that we can still compute the degree of entailment between any two sentences in the current setting. Sentence representations of our setting are either vectors (in the $\textmd{FbVect}_{\mathbb{R}}$ instantiation) or density matrices (in the  ${\cal CPM}(\textmd{FVect}_{\mathbb{R}})$ instantiation);  in each case one can calculate the representativeness of Shannon's entropy or the  KL divergence between them and compare the results in a case by case basis. What remains unproved is  that under which conditions these degrees remain nonzero, which is what is proved in Theorem   \ref{thm:main} for a special case.

KL-divergence and quantum relative entropy  give rise to an ordering on vectors and density matrices, respectively, which represents the difference in the information contents of the underlying words as given by vectors and density matrices. Exploring this order and the notion of logic that may arise from it is work in progress. The work of Widdows \cite{widdows} and Preller \cite{preller} might be relevant to this task. 

\section{Acknowledgements}

Sadrzadeh is supported by  EPSRC CAF grant EP/J002607/1 and Kartsaklis by  AFOSR grant FA9550-14-1-0079.  Balk{\i}r was supported by a Queen Mary Vice Principal scholarship, when contributing to this project.

\bibliographystyle{splncs03}
\bibliography{dih}

\end{document}